\documentclass{article}

\usepackage[nonatbib]{neurips_2019}

\usepackage[utf8]{inputenc} 
\usepackage[T1]{fontenc}    
\usepackage{hyperref}       
\usepackage{url}            
\usepackage{booktabs}       
\usepackage{amsfonts}       
\usepackage{nicefrac}       
\usepackage{microtype}      
\usepackage{xcolor}

\usepackage[round]{natbib}
\usepackage{color}
\usepackage{placeins}
\usepackage{graphicx}
\usepackage{subfigure}
\usepackage{bm}
\usepackage{booktabs}
\usepackage{threeparttable}
\usepackage{multirow}
\usepackage{amsmath}
\usepackage{algorithm}
\usepackage{algpseudocode}
\newtheorem{definition}{Definition}
\newtheorem{theorem}{Theorem}

\title{Attacking and Defending Deep Reinforcement Learning Policies}

\author{%
  Chao Wang \\
  \texttt{w-c15@tsinghua.org.cn} \\
}

\begin{document}

\maketitle

\begin{abstract}
Recent studies have shown that deep reinforcement learning (DRL) policies are vulnerable to adversarial attacks, which raise concerns about applications of DRL to safety-critical systems. In this work, we adopt a principled way and study the robustness of DRL policies to adversarial attacks from the perspective of robust optimization. Within the framework of robust optimization, optimal adversarial attacks are given by minimizing the expected return of the policy, and correspondingly a good defense mechanism should be realized by improving the worst-case performance of the policy. Considering that attackers generally have no access to the training environment, we propose a greedy attack algorithm, which tries to minimize the expected return of the policy without interacting with the environment, and a defense algorithm, which performs adversarial training in a max-min form. Experiments on Atari game environments show that our attack algorithm is more effective and leads to worse return of the policy than existing attack algorithms, and our defense algorithm yields policies more robust than existing defense methods to a range of adversarial attacks (including our proposed attack algorithm). 
\end{abstract}

\section{Introduction} 
Deep reinforcement learning (DRL) that utilizes deep neural networks (DNN) as function approximators has shown promising results in many sequential decision-making problems such as robotics control, computer games and autonomous driving (\citet{mnih2015human, bojarski2016end, levine2016end}). While DRL benefits from the strong representation ability of DNNs, it also inherits DNNs' vulnerabilities to adversarial attacks that small well-designed perturbations to inputs can cause wrong predictions (\citet{goodfellow2014explaining, fawzi2018analysis, kurakin2016adversarial,tramer2018ensemble}). The problem is even more serious in DRL; for example, an attack to a neural network-based module in a self-driving car may lead to an accident that happened in real life. 

A growing body of work is studying and improving the robustness of DRL to adversarial perturbations by proposing attack methods (\citet{huang2017adversarial, lin2017tactics, pattanaik2018robust}) and defense methods (\citet{behzadan2017whatever, mandlekar2017adversarially, pattanaik2018robust}). Most of these works directly borrow ideas from supervised learning (SL) to attack and defense DRL policies. For example, \citet{huang2017adversarial} propose to add perturbations to observations in order to decrease the probability of the best actions selected by the policy. \citet{mandlekar2017adversarially} implement adversarial training by modifying trajectory rollouts to include perturbations, which is similar to data-augmentation. However, DRL is different from SL, because there are no explicit labels telling whether performing a certain action at a certain state is \emph{right} or \emph{wrong} and the objective in DRL training is not minimizing a supervised loss. Therefore, we need a principled way to guide the attack and defense in DRL. 

In this work, we study the problem of attacking and defensing DRL and consider a general setting that policies are stochastic and known in advance to attackers. Since DRL derives a policy by maximizing its expected return, we argue that in the attack phase, attackers should target at minimizing the expected return. With this insight considering that training environments are generally unavailable to attackers, we use the policy's action-value function as a guide for efficient adversarial attacks. Specifically, perturbations are generated by minimizing the expectation of the action-value function with respect to the perturbed policy. The challenge is that the action-value function is unknow to attackers since the environments are unavailable to them. To address this challenge, we show that we can avoid using the action-value function; instead, we prove that perturbations can be generated in an equivalent manner by maximizing the cross entropy of the perturbed policy and the unperturbed policy. We propose an attack algorithm by maximizing the cross entropy of the two policies accordingly. As for the defense phase, by leveraging the concept of robust optimization, we formulate adversarial training of DRL policies as a max-min saddle point optimization problem, where the \emph{inner maximization} aims at finding perturbations that lead the policy to the worst-case scenario (which precisely matches the goal of adversarial attack), while the \emph{outer maximization} aims at finding policy parameters that maximize the expected return of the perturbed policy.  

The proposed attack and defense methods have two main merits. First, the attack algorithm generates perturbations that are certain to fool the agent to take the worst action at each state given that only the policy is available to attackers. Unlike some attack methods (\cite{pattanaik2018robust,lin2017tactics}) that need to access the action-value function, our method only requires knowing the policy but yields equivalent adversarial perturbations if the policy is optimal. Second, the robust optimization view of the defense algorithm guarantees that the learnt policy is robust to adversarial attacks. 

Our contributions are threefold. Firstly, we give a unified view on policy-based and value-based adversarial attacks in DRL and propose a practical attack algorithm that guarantees the agent is fooled to take the worst actions. Secondly, we formulate the defense problem as a max-min saddle point optimization problem by leveraging the robust optimization method. Thirdly, we propose a practical defense algorithm that is shown to be robust to a range of attacks (including our proposed attack) through extensive experiments.

\section{Related work}
Regarding adversarial attacks in DRL, \citet{kos2017delving} employ the same technique as \citet{huang2017adversarial} except that perturbations are applied only if the value function is over a threshold. Similarly, \citet{lin2017tactics} construct attacks by reducing the probability of the action with the highest action value and use a relative action preference function to guide the injection of adversarial perturbations. In contrast to \citet{lin2017tactics}'s work is the method that constructs adversaries by improving the probability of the action with the least action value (\citet{pattanaik2018robust}). \citet{mandlekar2017adversarially}, instead, build physically-plausible perturbations by maximizing the $\ell_2$ loss of the mean vector of the stochastic policy. Different from these methods, the attack algorithm presented in this paper is guaranteed to fool the agent to take the worst action at each state given that only the policy is available to attackers, as we will see in Section \ref{sec:attack}.

Only a few works focus on training DRL policies resistant to adversarial attacks. Different from \citet{mandlekar2017adversarially}'s work that implements adversarial training by including adversarial perturbations into unperturbed rollouts, \citet{behzadan2017whatever} demonstrate that under noncontiguous training-time attack, deep Q-network (DQN) agents with standard training techniques can recover and adapt to adversarial conditions. \citet{kos2017delving} present preliminary results showing that agents are able to be more resistant to adversarial attacks under retraining with perturbations. \citet{lin2017detecting} follow the idea that adversarial examples targeting at a neural network-based policy are not effective for action-dependent frame prediction models and thus use a learnt action-conditioned frame prediction module to detect adversarial attacks and to generate action proposals. Another class of relevant approaches (\citet{morimoto2005robust,pinto2017robust}) aim at training DRL agents robust to model uncertainties (changes in weight, friction or other parameters of dynamic systems) instead of adversarial attacks by formulating the problem as a two-player zero-sum game. \citet{pattanaik2018robust}, instead, solve the problem by leveraging adversarial attacks to fool the agent into sampling the worst trajectories and presenting a retraining scheme to train agents robust to model uncertainties. Though these defense algorithms have shown a certain degree of robustness to adversarial attacks, the performance of the perturbed policy is not guaranteed to be improved.

\section{Background}
\label{sec:background}
\subsection{Adversarial attacks}
In the context of SL, for any given sample $(x,y)$ and function $f$, the adversarial attack to $x$ under $f$ can be generated by solving the following constrained optimization problem:
\begin{equation}\label{equ:supervised_attack}
    \max_\delta L\left(f_{\theta}(x+\delta), y\right),\quad\mathrm{s.t.,} \quad \delta\in \mathcal{G},
\end{equation}
where $\theta$ is the parameter of function $f$, $L$ is the loss function, and $\mathcal{G}$ is a perturbation set, e.g., $\ell_\infty$ ball or $\ell_2$ ball around $x$ with $\epsilon$ radius (usually referred to as $\ell_{\infty}^{\epsilon}$ and $\ell_2^{\epsilon}$). Equation (\ref{equ:supervised_attack}) can be solved using the projected gradient descent (PGD) method (\citet{madry2018towards}). For example, for learning adversarial attacks in the $\ell_{\infty}^{\epsilon}$ setting, \citet{madry2018towards} perform PGD updates of the form:
\begin{equation}
    x_{t+1} = \text{Proj}_\infty^{x,\epsilon}\left(x_t + \alpha \cdot \mathrm{sgn}(\nabla L(f_{\theta}(x_t), y))\right). 
\end{equation}
Here function $\text{Proj}_\infty^{x,\epsilon}()$ operates as projecting the input onto the $\ell_\infty$ ball of radius $\epsilon$ around $x$ if the input is outside the ball, and $\alpha$ is the PGD step size.

\subsection{Reinforcement learning and policy gradient methods}
A RL problem is usually described as a Markov decision process (MDP). A MDP is defined by the tuple ($\mathcal{S}$, $\mathcal{A}$, $\mathcal{P}$, $r$, $\gamma$) (\citet{sutton2018reinforcement}), where $\mathcal{S}$ is the state space, $\mathcal{A}$ is the action space, $\mathcal{P}: \mathcal{S}\times \mathcal{A}\times \mathcal{S}\rightarrow \mathbb{R}$ is the state transition probability, $r: \mathcal{S}\times\mathcal{A} \rightarrow\mathbb{R}$ is the reward function, and $\gamma$ is the discount factor. The goal of reinforcement learning (RL) is to learn a parameterized control policy $a\sim\pi_\theta(\cdot|s)$ that maximizes the value function (or its surrogate):
\begin{equation}\label{equ:standard_loss}
    \max_\theta V_{\pi_{\theta}}(s_0), \quad \mathrm{where}\quad V_{\pi_\theta}(s_0)=\mathbb{E}_{s_{t\geq 1}\sim\mathcal{P}, a_{t\geq 0}\sim \pi_\theta} \left[\sum\limits_{t=0}^{\infty} \gamma^t r(s_t, a_t)\left|s_0, \pi_\theta\right.\right],
\end{equation}
where $s_0$ is the initial state. RL sometimes involves in estimating the action-value function
\begin{equation}\label{equ:critic_function}
    Q_{\pi}(s, a) = \mathbb{E}_{s_{t\geq 1}\sim\mathcal{P}, a_{t\geq 1}\sim \pi}\left[ \sum\limits_{t=0}^{\infty}\gamma^t r(s_t, a_t)\left|s_0 = a, a_0 = a, \pi\right.\right],
\end{equation}
which describes the expected return after taking action $a_t$ at state $s_t$ and thereafter following policy $\pi$.
By definition, the value function and the action-value function satisfy:
\begin{equation}\label{equ:relation_of_value_and_critic}
    V_\pi(s) = \sum_a \pi(a|s) Q_\pi(s,a).
\end{equation}
For ease of denotation, we focus on MDPs with discrete action spaces, but all algorithms and results can be directly applied to the continuous setting.

\section{Method}
In this section we first give a unified view on adversarial attacks and defenses in the context of DRL, and then introduce a practical attack algorithm and a practical defense algorithm.

\subsection{A unified view on attacks and defenses in DRL}
We solve the problem of adversarial attack and defense of DRL policies within the framework of robust optimization (\citet{madry2018towards,wald1945statistical}):
\begin{equation}\label{equ:min_max_loss}
    \max_\theta V_{\pi_{\theta}^{\delta}}(s_0), \quad \mathrm{where}\quad V_{\pi_\theta^{\delta}}(s_0) = \min_{\bm{\delta}}\mathbb{E}_{s_{t\geq 1}\sim\mathcal{P}, a_{t\geq 0}\sim \pi_\theta^\delta}\left[\sum\limits_{t=0}^{\infty} \gamma^t r(s_t, a_t)\left|s_0, \pi_\theta\right.\right].
\end{equation}
where $\pi_{\theta}^{\delta}$ refers to $\pi_{\theta}(a|s+\delta_s)$ and $\bm{\delta}$ denotes the set of adversarial perturbation sequences $\{\delta_{s_t}\}_{t\geq1}$ satisfying $\ell_{\infty}(\delta_{s_t})\leq \epsilon$ for all $t\geq 0$. 

This formulation provides us with a unified view on adversarial attacks and defenses in DRL. On one hand, the \emph{inner minimization} seeks to find a sequence of adversarial perturbations $\bm{\delta}$ that lead the current policy $\pi_\theta$ to the worst-case scenario. On the other hand, The \emph{outer maximization} aims at finding policy parameters $\theta$ that maximize the expected return of the perturbed policy, which is exactly the goal of training policies robust to adversarial attacks. 

\subsection{A practical attack algorithm}\label{sec:attack}


It is clear that the \emph{inner minimization} of Equation \eqref{equ:min_max_loss}, which generates adversarial perturbations, is another RL problem with infinity number of parameters $\bm{\delta}$. One would expect it intractable since learning optimal perturbations could be time-consuming, and worse still, the training environment is generally unavailable to attackers. Here we consider a practical setting that attackers greedily inject perturbations at different states. Unlike the SL setting in which attackers only need to fool classifiers to yield wrong labels (\citet{goodfellow2014explaining}), in the RL setting, the action-value function provides us with extra information that actions with smaller action values would result in lower expected returns. Accordingly, we define the optimal adversarial perturbations in DRL as below.

\begin{definition}\label{def:optimal_attack}
An optimal adversarial perturbation $\delta_s$ at state $s$ is any allowed perturbation that minimizes the expected return of the state, i.e., 
\begin{equation}\label{equ:critic_attack}
    \delta_s^* = \arg\min_{\delta_s} \sum_a \pi_\theta(a|s+\delta_s)Q_{\pi_\theta}(s,a), \quad\mathrm{s.t.,}\quad \ell_\infty(\delta_s) <= \epsilon.
\end{equation}
\end{definition}

If the minimum of Equation \eqref{equ:critic_attack} is tractable, it is certain that attackers fool the agent to select the worst action. Unfortunately, Equation \eqref{equ:critic_attack} is intractable since the action-value function is also unavailable to attackers.

The following theorem demonstrates that optimal perturbations can be generated in an equivalent manner without consulting the action-value function if the policy is optimal.

\begin{theorem}\label{the:theorem}
When the control policy $a\sim\pi(\cdot|s)$ is optimal, the action-value function and the policy satisfy the soft-max relation:
\begin{equation}\label{equ:themrem}
    \pi(a|s) \propto e^{ \frac{Q_{\pi}(s,a)}{\mu_s}}, \quad \mu_s>0,  \forall a\in \mathcal{A},
\end{equation}
where $C_s$ denotes the entropy of the policy, $\mu_s$ is a state-dependent constant, and as $C_s$ approaches zero, $\mu_s$ also approaches zero. Combining Equations \eqref{equ:critic_attack} and \eqref{equ:themrem}, we have:
\begin{equation}\label{equ:policy_attack}
    \delta_s^* = \arg\max_{\delta_s}\sum_a -\pi_\theta(a|s+\delta_s)\log\pi_{\theta}(a|s), \quad\mathrm{s.t.,}\quad \ell_\infty(\delta_s) <= \epsilon,
\end{equation}
\end{theorem}
\emph{Proof.} See Appendix \ref{appendixA}.

Theorem \ref{the:theorem} shows that optimal perturbations can be generated in an equivalent manner by maximizing the cross entropy of the perturbed policy and the unperturbed policy if the policy is optimal (which is often the case in the test phase). For ease of future discussion, this work will refer to Equation \eqref{equ:policy_attack} as \emph{policy attack}. Besides, given that PGD generates universal aversarial attacks (\citet{madry2018towards}), we use it to compute the optima of \emph{policy attack}. The attack algorithm is summarized in Algorithm \ref{alg:attack}.

\paragraph{Compare with previous attacks} Similar to \emph{policy attack} is the method that constructs adversaries by minimizing the cross entropy of the distribution that puts all weights on the action with the least action value and the perturbed policy (denoted as ``Min\_PGD'') (\citet{pattanaik2018robust}): $\delta_s^* = \arg\min_{\delta_s} -\log\pi_\theta(a^w|s+\delta_s)$, where $a^w$ denotes the action with the lowest action value.
\emph{Policy attack} deviates from Min\_PGD in two aspects: firstly, \emph{policy attack} does not require accessing the action-value function; secondly, \emph{policy attack} degenerates into Min\_PGD if the policy becomes deterministic since as $C_s$ approaches zero, the soft-max relation degenerates into arg-max relation. Another similar attack method generates adversaries by minimizing the probability of the most likely selected action (\citet{huang2017adversarial})(denoted as ``Max\_PGD''): $\delta_s^* = \arg\max_{\delta_s} -\log\pi_\theta(a^*|s+\delta_s)$, where $a^*$ denotes the action with the highest probability. Different from \emph{policy attack}, Max\_PGD borrows ideas from SL that attackers fool classifiers by preventing it from selecting \emph{right} labels.

\begin{algorithm}
\caption{Adversarial Attack with PGD}
\label{alg:attack}
\begin{algorithmic}[1]
\Procedure{Attack}{$\pi_\theta(a|s)$ $\epsilon$, $n$, $\alpha$}\quad $\triangleright$ Adversarial attack function takes current policy $\pi_\theta(a|s)$, adversarial attack magnitude constraint $\epsilon$,  number of PGD steps $n$ and step size $\alpha$ as input
    \State Randomly initialize the perturbed observation $x_0 = s + \delta_0$ s.t. $\ell_\infty(\delta_0)\leq\epsilon$
    \For {$i=0, \cdots, n-1$}
        \State Prepare the loss function $L(x) = \sum\limits_a \pi_\theta(a|x)\log\pi_\theta(a|s)$
        \State Update state $x$ by $x_{i+1}$ = $\text{Proj}_\infty^{x,\epsilon}\left(x_i + \alpha \cdot \mathrm{sgn}\left(\left.\nabla_x L(x)\right|_{x=s + \delta_i}\right)\right)$
        \State Update adversarial attack $\delta_{i+1} = x_{i+1}-s$
	\EndFor
	\State Return $\delta_{n-1}$
\EndProcedure
\end{algorithmic}
\end{algorithm}

\begin{algorithm}
\caption{Adversarial Training with Policy Attack (ATPA)}
\label{alg:defense_policy}
\begin{algorithmic}[1]
\State Randomly initialize policy network $\pi_\theta(a|s)$ and action-value network $Q_\omega(s,a)$ of the attacked policy $\pi_{\theta}(a|s+\delta_s)$ with weights $\theta$, and $\omega$
\State Set adversarial attack magnitude constraint $\epsilon$, PGD steps $n$ and step size $\alpha$
\For {$\mathrm{iteration}=1, \cdots$}
    \State Rollout a trajectory $\tau_{\pi_\theta^\delta}=(s_0, a_0, r_0, \cdots, s_{T-1}, a_{T-1}, r_{T-1})$ by running policy $\pi_\theta(a|s+\delta_s)$ for $T$ time steps, where $\delta_s$ is obtained by running $\mathrm{\textsc{Attack}}(\pi_\theta(a|s), \epsilon, n, \alpha)$
    \State Update action-value network weights $\omega$ using trajectory $\tau_{\pi_\theta^\delta}$
    \State Optimize Equation \eqref{equ:min_max_loss} using policy gradient methods (e.g., A3C, PPO), trajectory $\tau_{\pi_\theta^\delta}$, action values $\{Q_w(s_t,a_t)\}_{t=0}^{T-1}$ and policy probabilities $\{\pi_\theta(a_t|s_t+\delta_{s_t})\}_{t=0}^{T-1}$
\EndFor
\end{algorithmic}
\end{algorithm}

\subsection{A practical defense algorithm}
Within the robust optimization framework, defense against adversarial perturbations can be realized by improving the performance of the perturbed policy in the worst-case scenario. Therefore, during the training phase, the perturbed policy $\pi_\theta^\delta$ is used to interact with the environment and in the mean time the action-value function $Q_{\pi_\theta^\delta}(s,a)$ (or the value function $V_{\pi_\theta^\delta}(s)$) of the perturbed policy is estimated to assist policy training.

Our proposed defense algorithm, adversarial training with \emph{policy attack} (ATPA), is summarized in Algorithm \ref{alg:defense_policy}. Specifically, we still use \emph{policy attack} to generate perturbations in the training phase even though the value function is no longer guaranteed to be reduced (since the policy is not optimal). The intuition behind is that though in the early training stage, the policy might be unrelated to the action-value function, as training progresses, they should gradually meet with the soft-max relation. On the other hand, if we use Equation (\ref{equ:critic_attack}) to generate adversaries, we need to explicitly estimate the action-value function $Q_{\pi_\theta}(s,a)$ of the unperturbed policy $\pi_\theta(a|s)$. One would expect it intractable since trajectories are collected by running the perturbed policy and estimating the action-value function of the unperturbed policy using these data could be highly inaccurate.

\section{Experiments}
\label{sec:experiments}
The effectiveness of our proposed methods is evaluated in several Arcade learning environments (\verb+Boxing+, \verb+SpaceInvaders+ and \verb+Carnival+) using PPO. Implementation of PPO is based on OpenAI baselines (\citet{baselines}). Details and the hyperparameter configuration can be found in Appendix \ref{appendixB}. All adversarial perturbations are obtained using PGD with $\epsilon=4$ and $n=7$.

Our experiments aim at answering the following questions: 1) Is \emph{policy attack} more efficient than existing attack algorithms? 2) Will the defense algorithm yield more robust policies? 3) What challenges adversarial perturbations bring to DRL algorithms? To answer these questions, we compare our attack/defense algorithms with several existing attack and defense algorithms.

The perturbations we consider are: 1) Max\_PGD (\citet{huang2017adversarial}); 2) Min\_PGD (\citet{pattanaik2018robust}); 3) Perturbations uniformly randomly selected from the feasible set, denoted by ``Random''. Policies are also evaluated in the perturbation-free setting, denoted by ``No Attack''. Our \emph{policy attack} is denoted by ``CE\_PGD''. Besides, while the policy and the action-value function satisfy the soft-max relation, we implement Min\_PGD in the test phase without consulting the action-value function. 

The defense algorithms we consider are: 1) Pretraining without adversarial attacks and retraining with adversarial attacks (\citet{kos2017delving}), denoted by ``StageWise''. In the retraining stage, only a portion of states are attacked. 2) Training with dynamic data augmentation (\citet{mandlekar2017adversarially}), denoted by ``DataAugment''. The algorithm aims at training policies robust to both adversarial perturbations and model uncertainties. For ease of comparison, we reimplement StageWise and DataAugment using PPO. For StageWise, we use the first third training steps for pretraining and add perturbations to all states in the retraining stage. For DataAugment, we follow the curriculum learning approach (\citet{mandlekar2017adversarially}) to train agents on increasing perturbation frequency $\phi$ and set $\phi_{max} =0.5$. All other hyperparameters are the same as those of ATPA.

\subsection{Evaluating attack algorithms}
\begin{figure}
\centering
\subfigure{
\includegraphics[width=3.73cm]{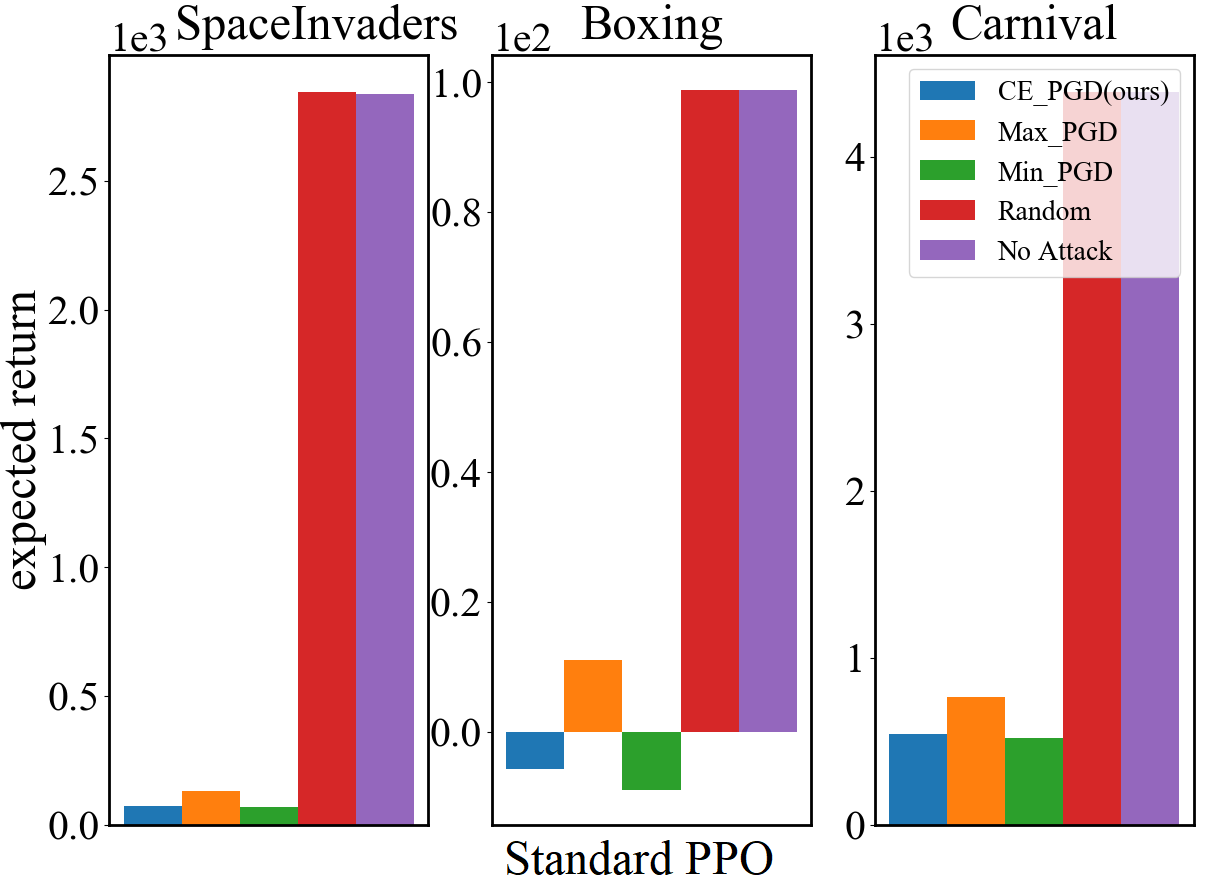}}
\subfigure{
\includegraphics[width=3.1cm]{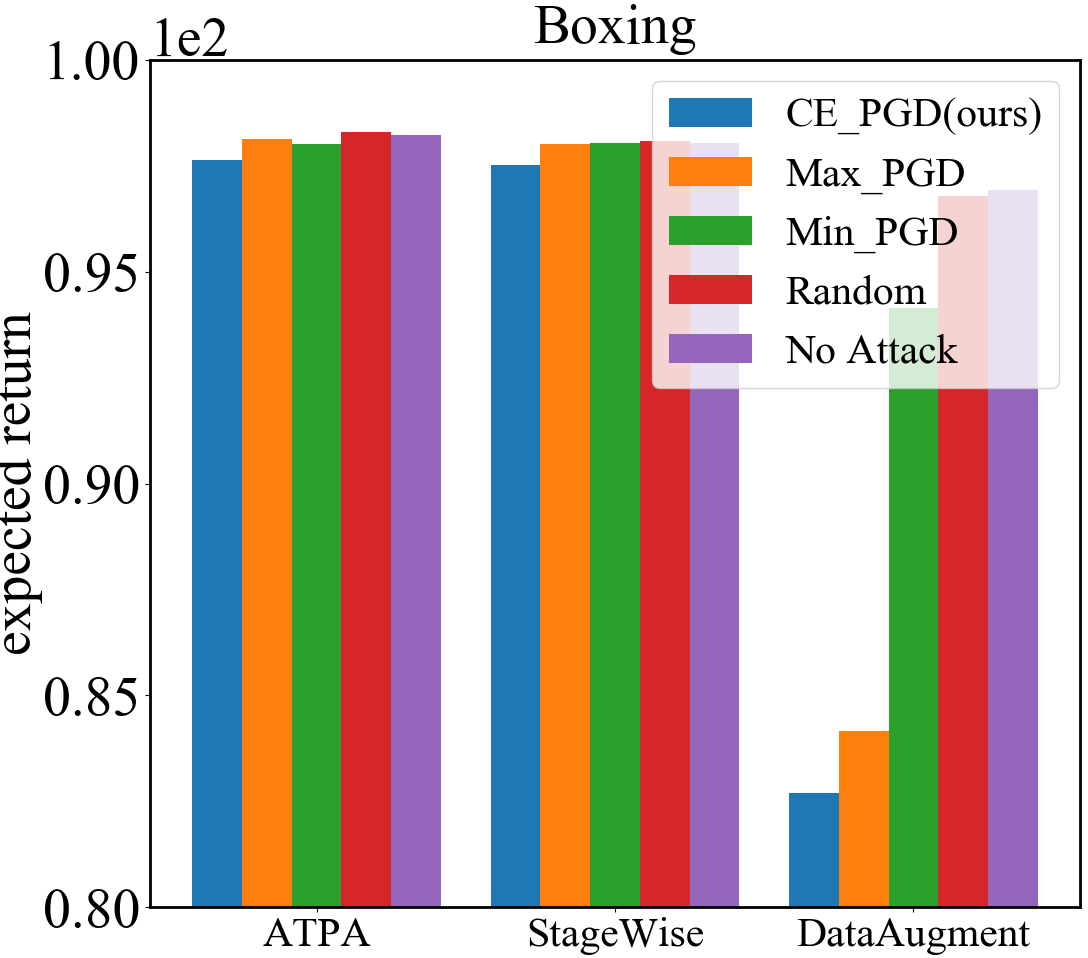}}
\subfigure{
\includegraphics[width=3.1cm]{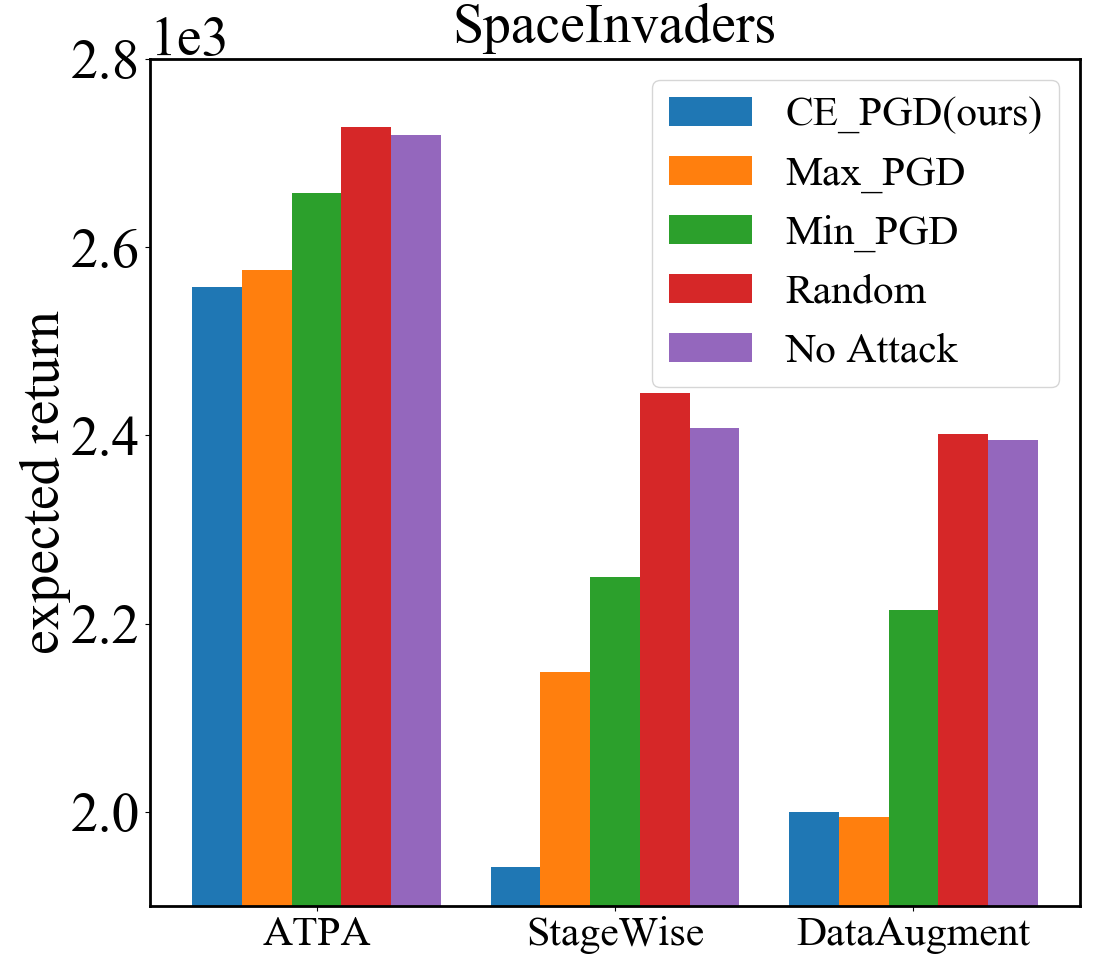}}
\subfigure{
\includegraphics[width=3.1cm]{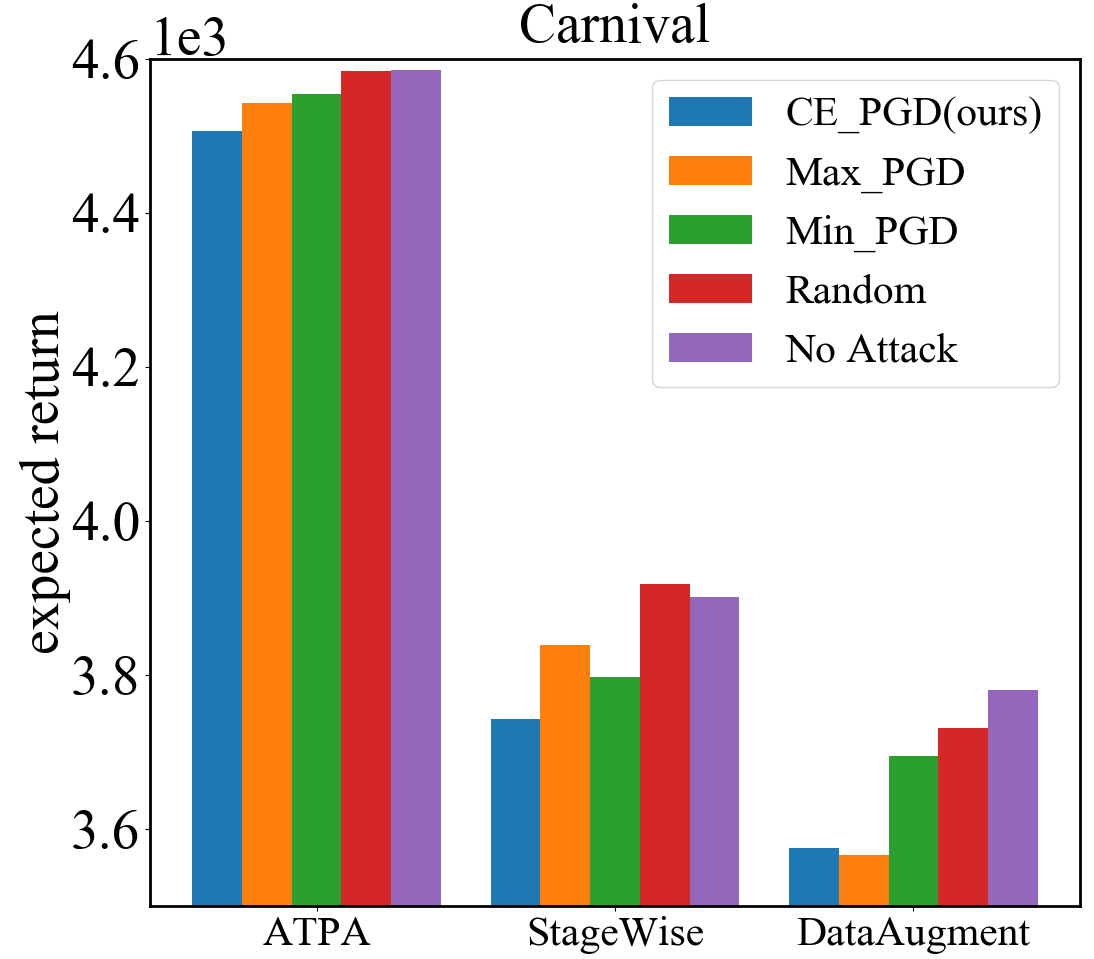}}
\caption{Expected returns of policies corrupted with adversarial perturbations generated by different attack algorithms. Each return is obtain by averaging rewards of $2500$ episodes ($5$ models (trained with different seeds) $\times 10$ environment seeds for each model $\times$ $50$ episodes for each seed).}\label{fig:attack_models}
\end{figure}

Figure \ref{fig:attack_models} illustrates the expected returns of different policies corrupted with different perturbations. As shown in the first subfigure, though policies trained with standard PPO yield the highest scores in all cases, adversarial perturbations almost disable them. The result demonstrates that DRL policies without adversarial training are inherently vulnerable to adversarial attacks. Besides, as standard PPO is unable to resist against adversarial perturbations, all the three adversarial attack algorithms are efficient enough to disable standard policies.

The three subfigures in the right hand side show the results of different perturbations attacking adversarially trained policies. Interestingly, both adversarially trained policies and standard policies are resistant to random perturbations. On the contrary, adversarial attacks degenerate the performance of different policies. The result depends on tested environments and defense algorithms. It seems that in the simple \verb+Boxing+ environment, both ATPA and StageWise have derived the ability against adversarial perturbations. Therefore, there is only a small performance gap between the three adversarial attack algorithms. By contrast, in the relatively harder \verb+SpaceInvaders+ environment, policies perturbed by our \emph{policy attack} algorithm yield significantly lower returns. Overall, our proposed \emph{policy attack} algorithm yields the lowest returns in most cases, demonstrating that it is indeed the most efficient among all tested adversarial attack algorithms.

\subsection{Evaluating defense algorithms}
\begin{figure}
\centering
\subfigure{
\includegraphics[width=4.05cm]{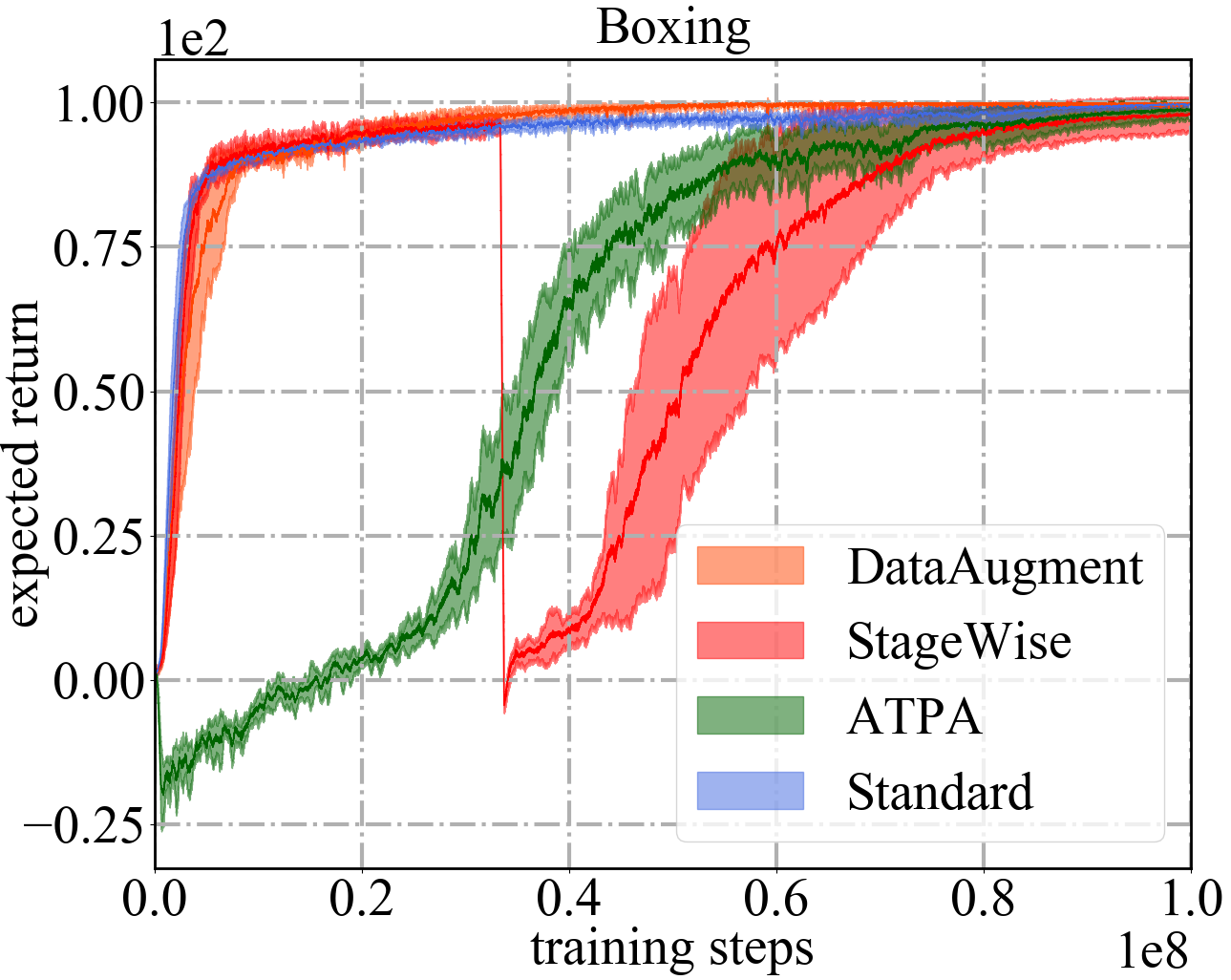}}%
\subfigure{
\includegraphics[width=3.87cm]{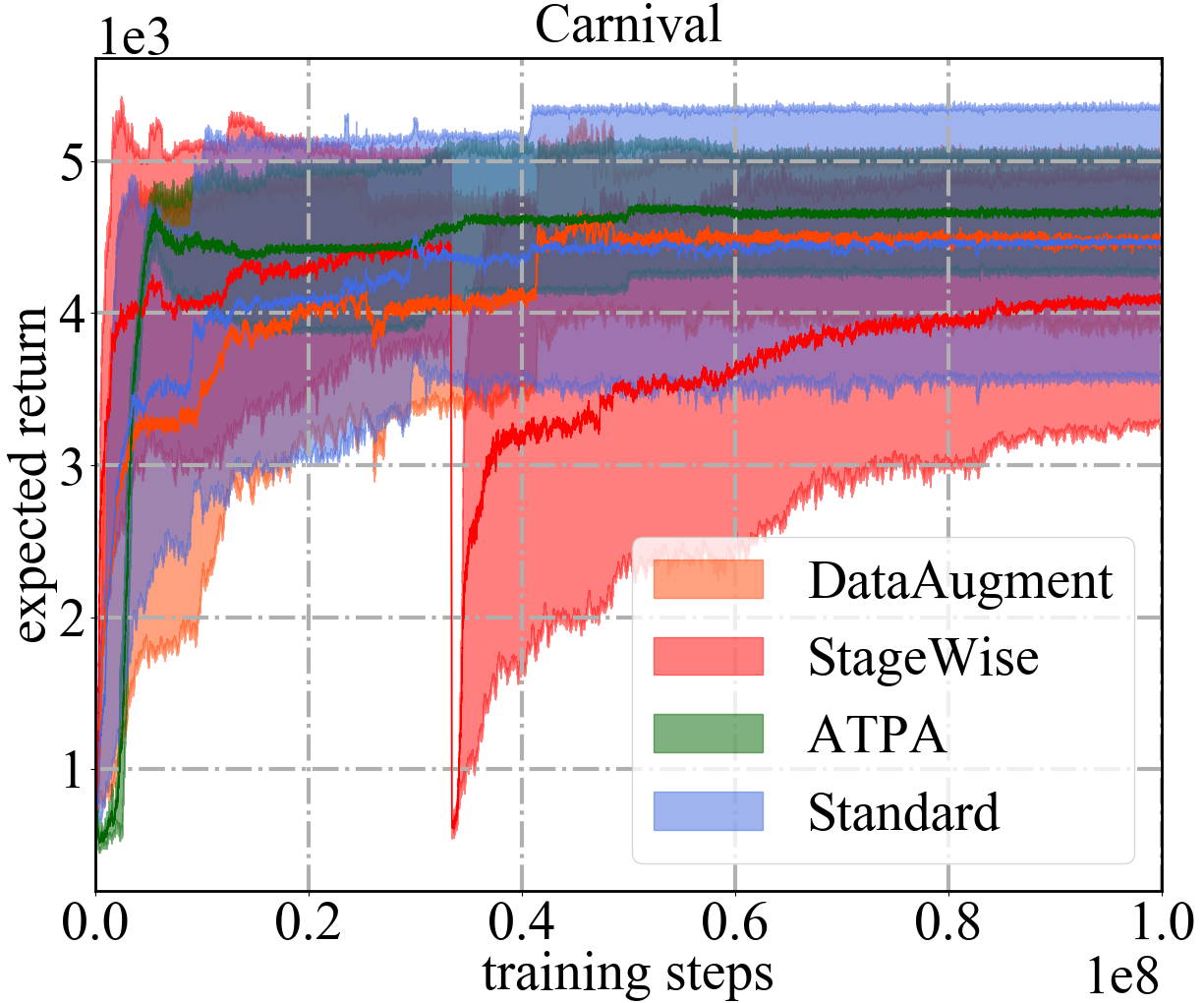}}%
\subfigure{
\includegraphics[width=4.0cm]{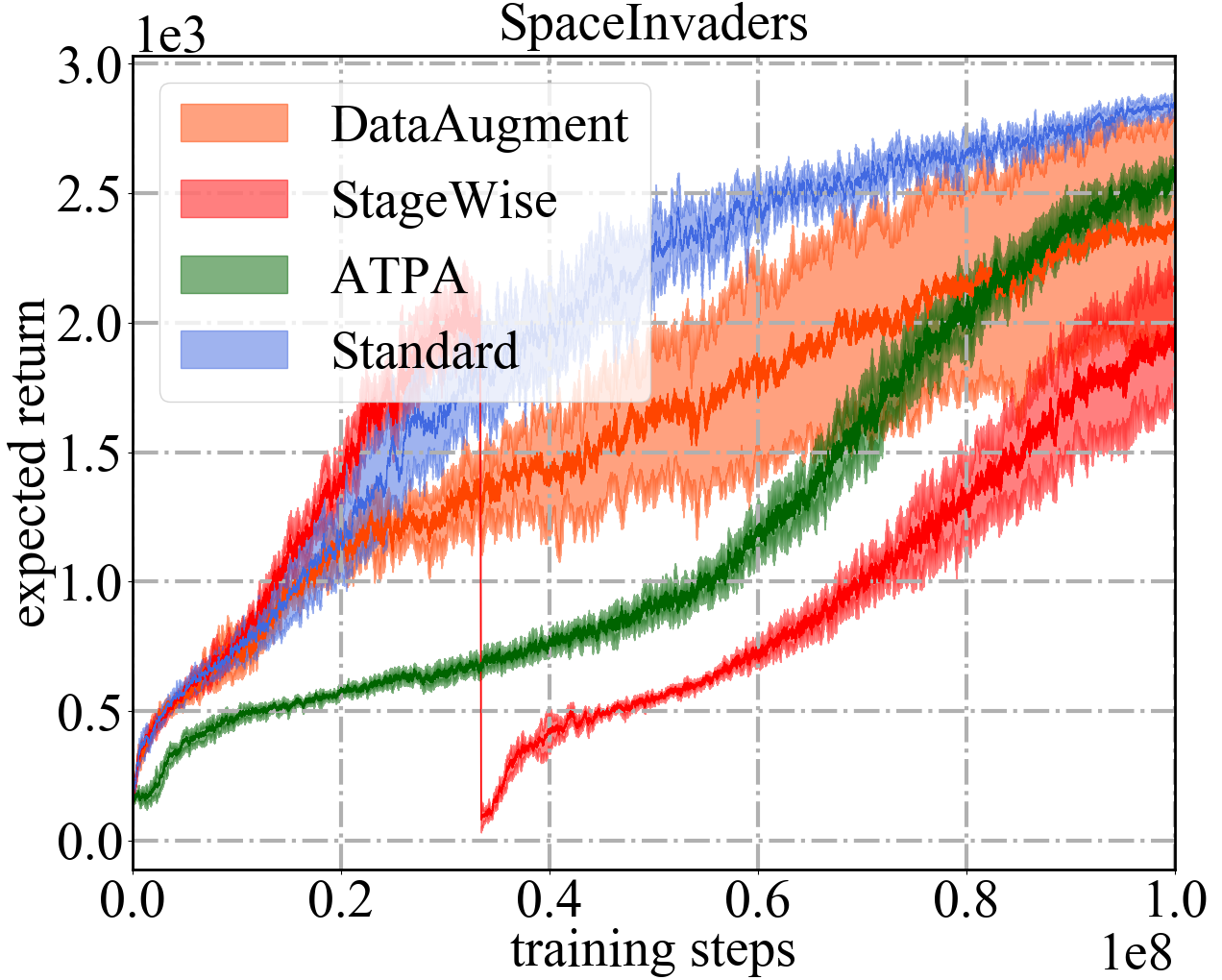}}%
\caption{Learning curves of different training algorithms. Each algorithm is trained with $5$ different seeds. Since different policies are used to interact with the environments (e.g., ATPA uses the perturbed policy while DataAugment uses the perturbation-free policy), the performance curves only represent expected returns of the policies used for environment interaction.}\label{fig:convergence_curves}
\end{figure}

\begin{table}
\centering
\caption{Expected returns of different policies corrupted with different perturbations. Each return is obtain by averaging rewards of $2500$ episodes ($5$ models (trained with different seeds) $\times 10$ environment seeds for each model $\times$ $50$ episodes for each seed).}\label{tab:different_attack}
    \begin{tabular}{l|l|cccc}
        \toprule 
        \multirow{2}{*}{Environments} & \multirow{2}{*}{Perturbations} & \multicolumn{4}{c}{Training algorithms}\\
        & & Standard & ATPA(Ours) & StageWise & DataAugment\\
        \hline
        \multirow{5}{*}{Boxing}        & CE\_PGD(Ours)              & -5.65                 & \textbf{97.63}            & 97.52             & 82.69\\
                                       & Max\_PGD                   & 11.15                 & \textbf{98.13}            & 98.02             & 84.15 \\
                                       & Min\_PGD                   & -8.91                 & \textbf{98.03}            & 98.03             & 94.14\\
                                       & Random                     & \textbf{98.70}        & 98.30                     & 98.08             & 96.78\\
                                       & No Attack                  & \textbf{99.44}        & 98.23                     & 98.04             & 96.92\\
                                       
        \hline
        \multirow{5}{*}{SpaceInvaders}   
                                       & CE\_PGD(Ours)              & 63.31                 & \textbf{2557.98}          & 1974.53           & 2000.16\\
                                       & Max\_PGD                   & 132.70                & \textbf{2575.79}          & 2148.43           & 1994.72\\
                                       & Min\_PGD                   & 68.36                 & \textbf{2658.02}          & 2249.77           & 2214.56\\
                                       & Random                     & \textbf{2845.76}      & 2727.64                   & 2444.70           & 2401.38\\
                                       & No Attack                  & \textbf{2836.46}      & 2718.72                   & 2407.65           & 2394.75\\
                                       
        \hline
        \multirow{5}{*}{Carnival}      & CE\_PGD(Ours)              & 542.36                & \textbf{4506.39}          & 3743.26           & 3574.90\\
                                       & Max\_PGD                   & 765.52                & \textbf{4542.42}          & 3838.80           & 3566.73\\
                                       & Min\_PGD                   & 523.50                & \textbf{4554.94}          & 3798.00           & 3694.41\\
                                       & Random                     & 4388.75               & \textbf{4585.01}          & 3918.65           & 3731.10\\
                                       & No Attack                  & 4386.52               & \textbf{4586.36}          & 3918.20           & 3780.79\\

        \bottomrule
    \end{tabular}
\end{table}

Learning curves of different defense algorithms along with standard PPO appear in Figure \ref{fig:convergence_curves}. Note that the performance curves only represent the expected returns of the policies that are used to interact with the environment. (e.g., DataAugment uses the perturbation-free policy while ATPA uses the perturbed policy). Of all training algorithms, our proposed ATPA exhibits the lowest training variance, and thus is more stable than other algorithms. We also notice that ATPA progresses much slower than standard PPO, especially in the earlier training stage. That contributes to the fact that in the earlier training stage, adversarially perturbed policies are highly unstable (we postpone the explanation to the next section).
\begin{figure}
\centering
\subfigure{
\includegraphics[width=4.0cm]{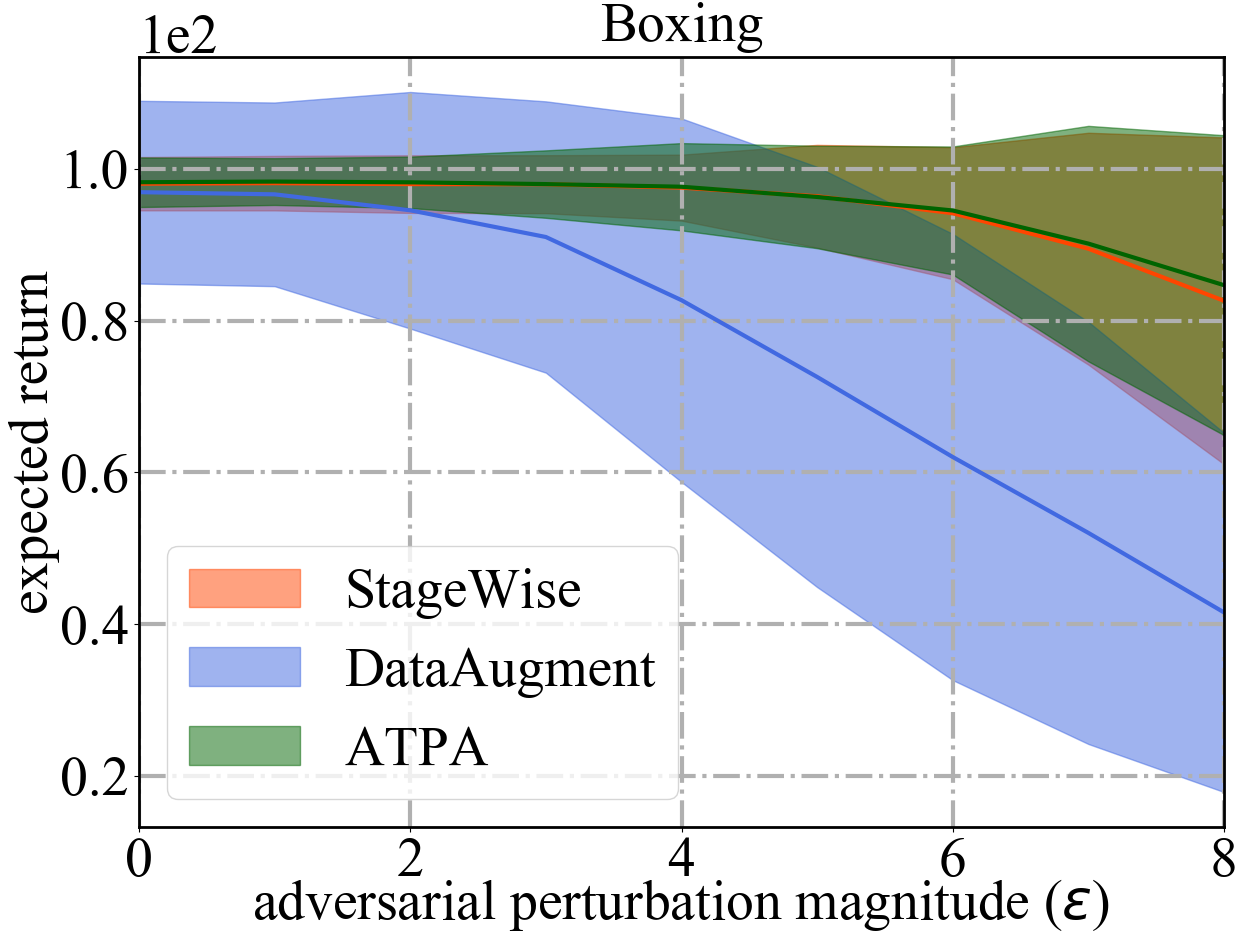}}
\subfigure{
\includegraphics[width=3.85cm]{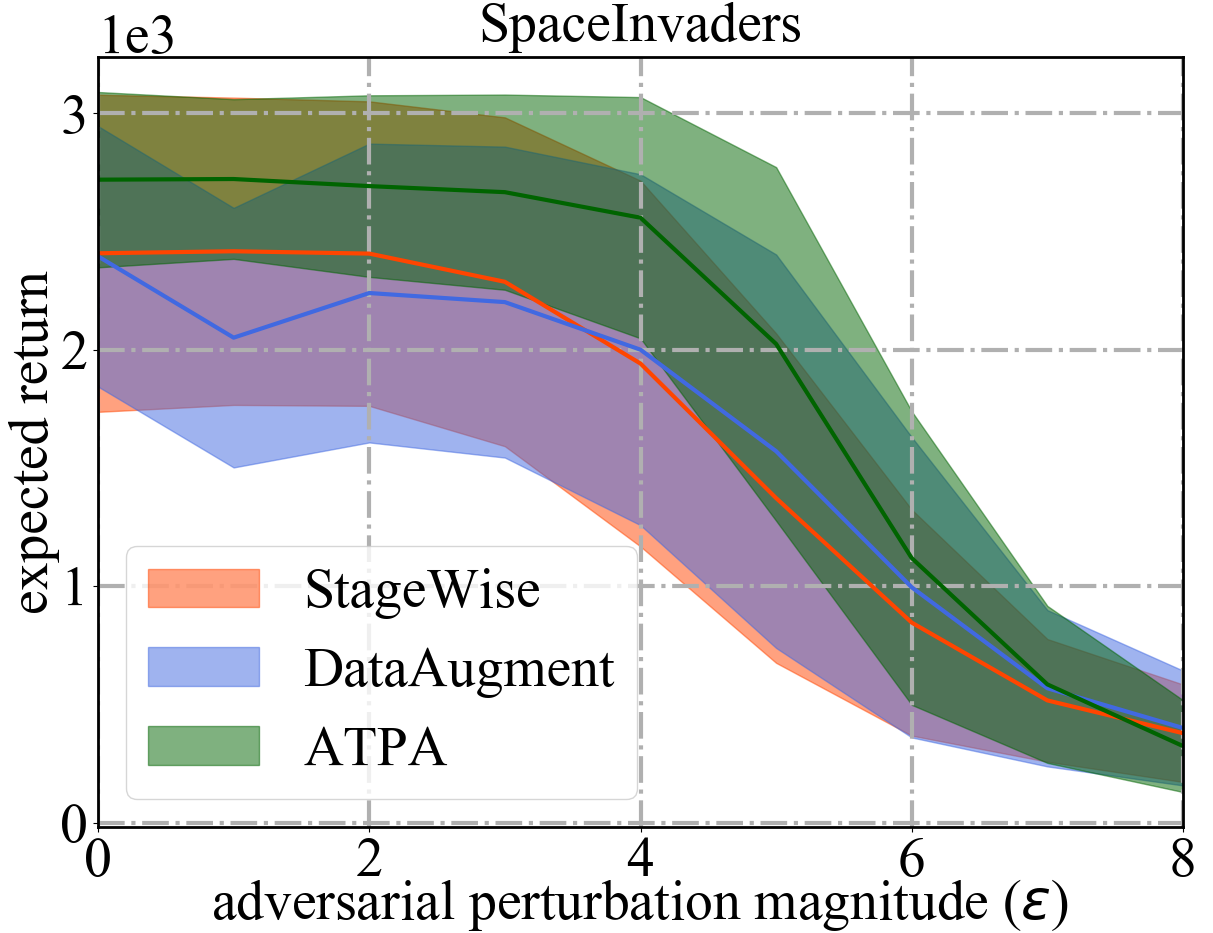}}
\subfigure{
\includegraphics[width=3.85cm]{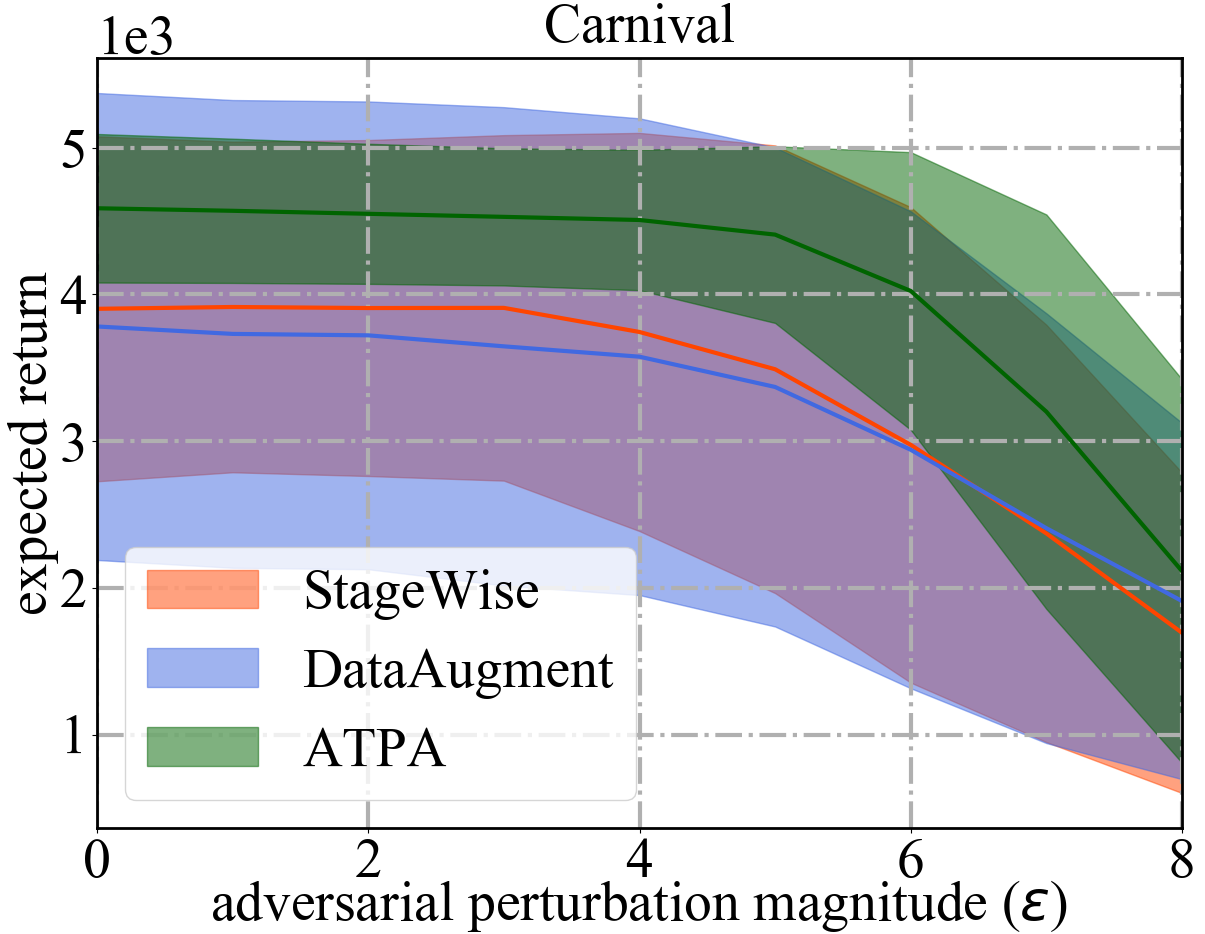}}
\caption{Robustness of policies to different magnitudes of adversarial perturbations. For each $\epsilon$, we compute the mean and variance of rewards of $2500$ episodes ($5$ models (trained with $5$ different seeds) $\times 10$ environment seeds for each model $\times 50$ episodes for each seed).}\label{fig:epsilon_curves}
\end{figure}

Table \ref{tab:different_attack} summarizes the test-time expected returns of policies trained with different algorithms and corrupted with different perturbations. Policies trained with ATPA retain resistant to all kinds of adversarial perturbations. In contrast, though StageWise and DataAugment learn to handle adversarial attacks to some extent, they are less efficient than ATPA in all cases. Besides, DataAugment seems to overfit the training environment. For example, in the \verb+Carnival+ environment, the expected return of the policy trained with DataAugment has reached $4500$ (shown in Figure \ref{fig:convergence_curves}), but its test-time return is less than $3800$ (in the perturbation-free situation).

For a broader comparison, we evaluate the robustness of these defense algorithms to different magnitudes of adversarial perturbations generated by the most efficient \emph{policy attack} algorithm. The result appears in Figure \ref{fig:epsilon_curves}. ATPA again achieves the highest scores in all cases. Additionally, the evaluation variance of ATPA is much smaller than that of StageWise and DataAugment, demonstrating that ATPA has even stronger generation ability.
 
\subsection{The landscape of adversarial perturbations}\label{sec:landscape}
\begin{figure}
\centering
\subfigure{
\includegraphics[width=4.0cm]{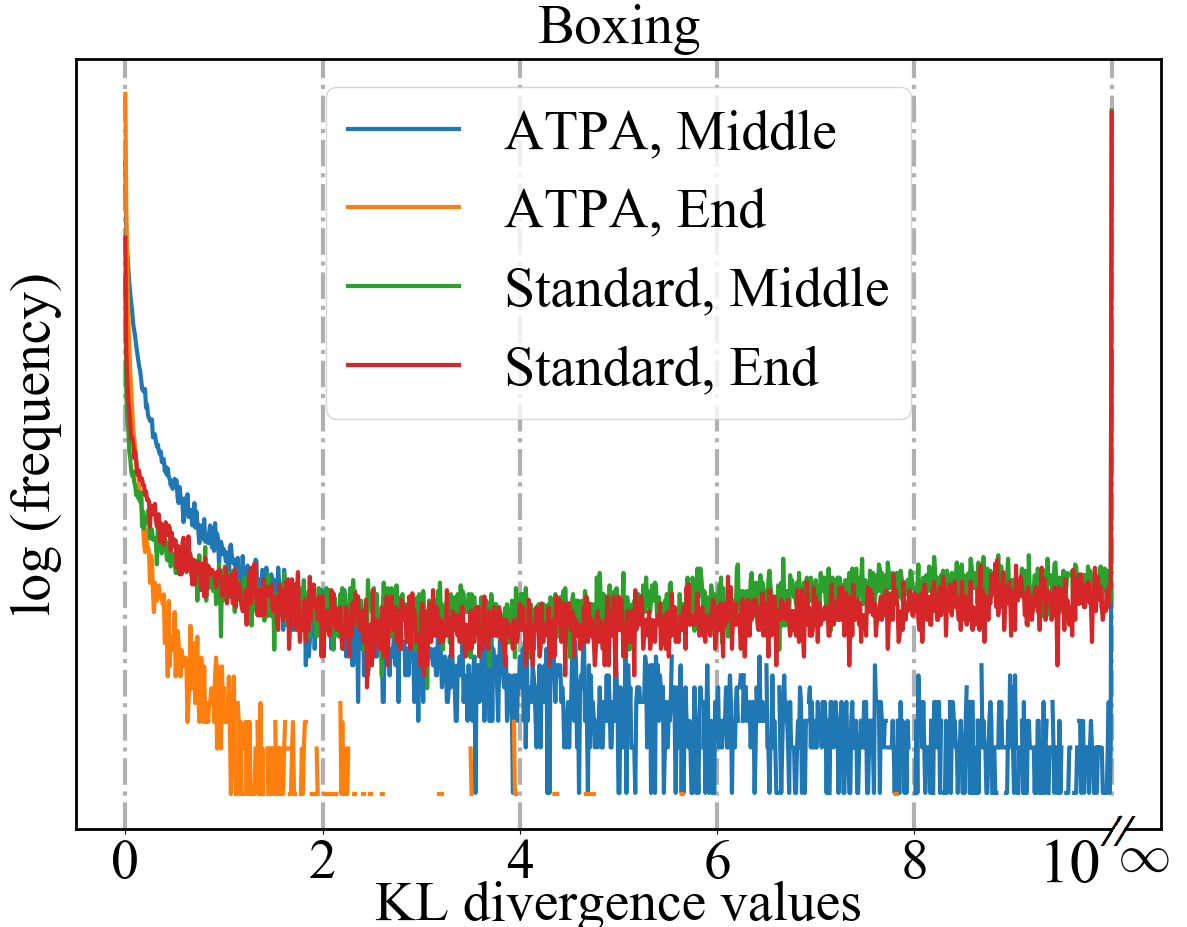}}%
\subfigure{
\includegraphics[width=4.0cm]{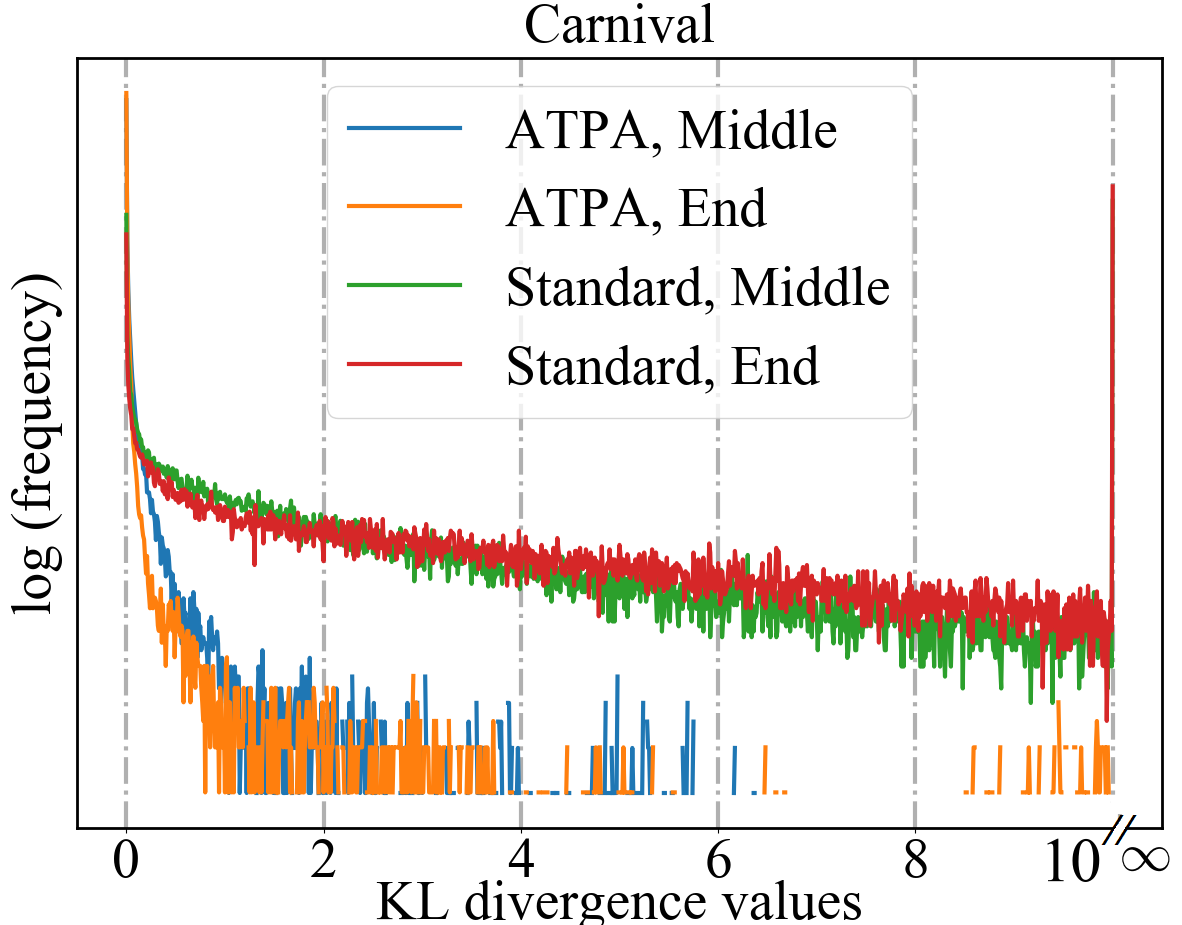}}%
\subfigure{
\includegraphics[width=4.0cm]{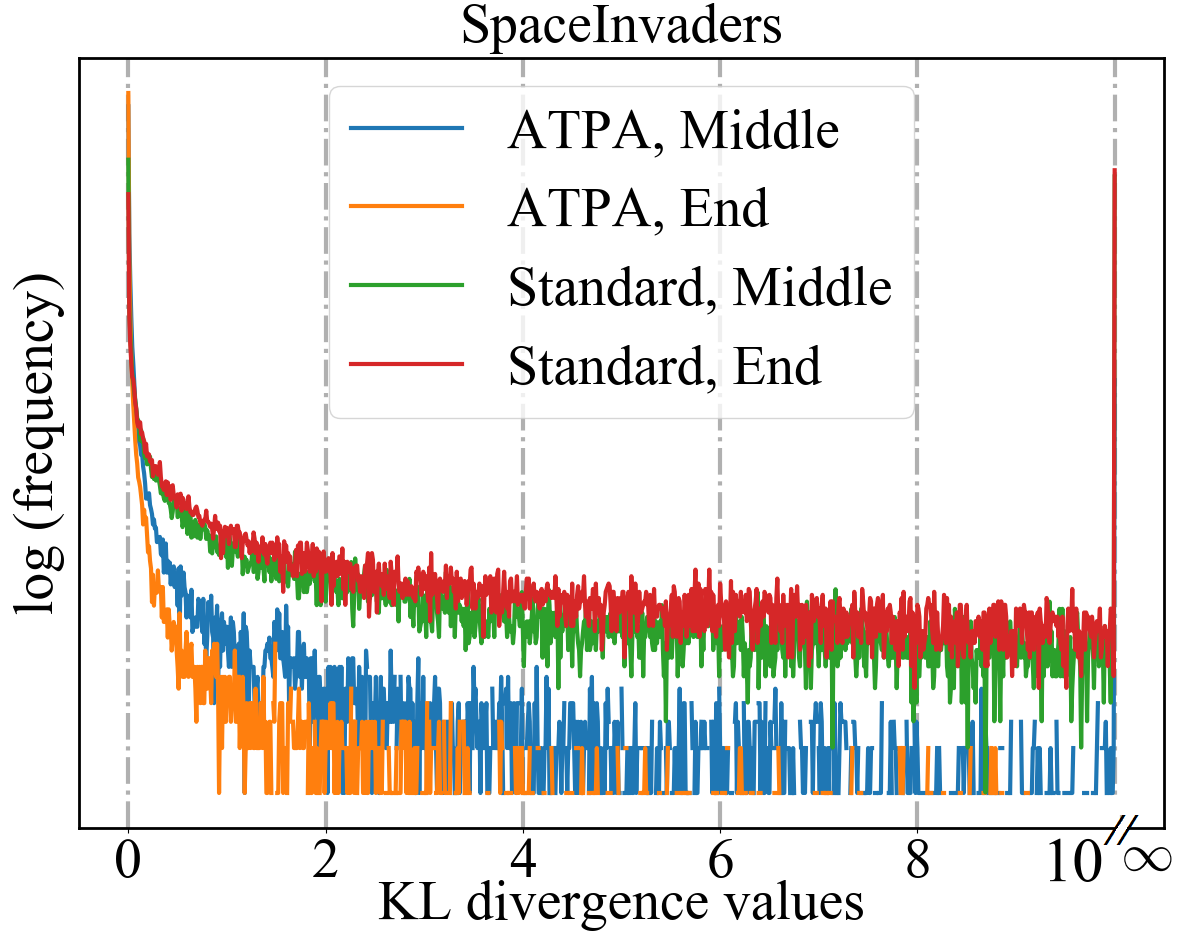}}%
\caption{Distributions of maximum KL divergence values (in different training stages (Middle, End)) of perturbed policies given by optimizing Equation \eqref{equ:policy_attack} with different initial points. In each environment, $50,000$ randomly selected states are used to obtain the distribution of maximum KL divergence values. For each state, we perform projected gradient descent (PGD) from $50$ random points in the $\ell_\infty^\epsilon$ ball and iterate PGD for $7$ steps. The maximum value for each state is obtained by taking the maximum of $2500$ KL divergence values of the $50$ perturbed policies.}\label{fig:divergence_statistics}
\end{figure}

To reach similar performance, ATPA requires more training data than standard PPO algorithm does. We dig into the problem by investigating the stability of perturbed policies. In particular, we calculate the Kullback–Leibler (KL) divergence (a measure of similarity between two probability distributions) values of perturbed policies obtained by performing \emph{policy attack} using PGD with different random initial points in the middle and at the end of the training process. Empirical distributions of maxima of KL divergence values appear in Figure \ref{fig:divergence_statistics} (illustrations of KL divergence maps of several randomly sampled states in the \verb+Boxing+ environment can be found in Figure (\ref{fig:divergence_map}) in Appendix \ref{appendixC}). As observed, without adversarial training, we constantly observe large KL divergence values even though standard PPO has converged (which indicates that policies are highly unstable to perturbations given by performing PGD with different initial points). By contrast, though different random starts may yield quite different perturbed policies in the middle training stage, large KL divergence values are barely observed when ATPA converges. The instability of policies to random starts may cause training instability that even policy parameters $\theta_t$ remain unchanged, the policy at two contiguous training steps, say, $\pi_{\theta_t}(a|s+\delta_s)$ and  $\pi_{\theta_{t+1}}(a|s+\delta_s)$, may still deviate a lot from each other. While DRL algorithms such as PPO and TRPO require that policies are improved gradually (for algorithm convergence and for ease of value estimation), the instability of perturbed policies partially explains why ATPA converges slower than standard PPO.

We also visualize adversarial perturbations in different environments (illustrations appear in Figure \ref{fig:adversarial_attacks} in Appendix \ref{appendixC}). Interestingly, adversarial perturbations are intrinsically different from random noise and are explainable after all. For example, in the \verb+SpaceInvaders+ environment, players manipulate a laser shooter and fire at descending aliens. Attackers fool a player by adding perturbations to pixels that represent the laser shooter and aliens to prevent the player from correctly recognizing them. The interesting discovery indicates that DRL algorithms indeed learn to play games by discovering controllable elements of the observations rather than simply overfitting environments. It also inspires us that by introducing attention-aware or contingency-aware mechanisms (\citet{rao2017attention, choi2018contingencyaware}), adversarially trained policies may be more robust to adversarial perturbations.
\thanks{}

\section{Conclusion}
In this work, by leveraging the robust optimization method, we formulate the problem of training DRL policies robust to adversarial perturbations as a max-min saddle point optimization problem. Within the framework of robust optimization, a practical attack algorithm, say, \emph{policy attack}, is proposed to generates adversarial attacks that lead the policy to the worst-case scenario. Correspondingly, a defense algorithm, say ATPA, is proposed to improve the worst-case performance of the policy. Extensive experiments demonstrate that our attack algorithm yields the most efficient adversarial perturbations and the defense algorithm yields reliable policies robust to a range of adversarial attacks. For future study, we think the combination of adversarial training with attention-aware or contingency-aware mechanisms can lead to further improvement on robustness of DRL policies to adversarial perturbations.

\newpage
\bibliography{bibliography}
\newpage
\appendix
\section{Proof of Theorem \ref{the:theorem}}\label{appendixA}
When the stochastic policy $\pi(a|s)$ is optimal, the value function $V_{\pi}(s)$ also reaches its optimum, that is to say, at each state $s$, we cannot find another action distribution that further increases $V_{\pi} (s)$. Accordingly, given the optimal action-value function $Q_\pi(s,a)$, we can recover the optimal policy $\pi(a|s)$ by solving the following constrained optimization problem:
\begin{equation}\label{equ:constrained}
\begin{aligned}
    \max_{\pi(a|s), a\in \mathcal{A}} \sum_{a\in \mathcal{A}}\pi(a|s)Q_\pi(s,a)\\
    \mathrm{s.t.,} \sum_{a\in\mathcal{A}} \pi(a|s) - 1=0\\
    \pi(a|s) \geq 0, ~a\in\mathcal{A}\\
    -\sum_{a\in\mathcal{A}}\pi(a|s)\log(\pi(a|s))-C_s\geq 0,\\
\end{aligned}
\end{equation}
where the second and the third rows encode that policy $\pi$ is a probability distribution, and the last row encodes that policy $\pi$ is a stochastic policy (so its entropy must be larger than a state-dependent constant $C_s$). According to the Karush-Kuhn-Tucker condition (\citet{kuhn1951}), solving Equation \eqref{equ:constrained} is equivalent to solving

\begin{equation}\label{equ:dual}
\begin{aligned}
    L(\pi, \lambda, \boldsymbol{\mu}) = 
    -\sum\limits_{a\in\mathcal{A}} \pi(a|s)Q_\pi(s,a) + \lambda \left(\sum\limits_{a\in\mathcal{A}} \pi(a|s) -1\right)\\ - \sum_{a\in \mathcal{A}}\mu_a\pi(a|s) 
    + \mu_s\left(C_s + \sum_{a\in \mathcal{A}}\pi(a|s)\log\pi(a|s)\right)\\
    \nabla_\pi L(\pi, \lambda, \boldsymbol{\mu}) = 0\\
    \sum_{a\in\mathcal{A}}\pi(a|s) - 1 = 0\\
    \mu_a\pi(a|s)=0, ~\forall a\in\mathcal{A}\\
    \mu_s\left(C_s + \sum_{a\in \mathcal{A}}\pi(a|s)\log\pi(a|s)\right)=0\\
     \lambda \neq 0\\
     \mu_s \geq 0, \mu_a \geq 0, ~\forall a\in\mathcal{A},\\
\end{aligned}
\end{equation}
where $\boldsymbol{\mu} = \{\mu_a\}_{a\in\mathcal{A}} \cup \{\mu_s\}$.
Assuming that $\pi(a|s)$ is positive for all actions $a \in \mathcal{A}$ (then $\mu_a=0$ for $\forall a\in\mathcal{A}$), 
the second equality gives:
\begin{equation}
    -Q_\pi(s,a) + \lambda + \mu_s(\log\pi(a|s) + 1) = 0.
\end{equation}
It is certain that $\mu_s > 0$ because if $\mu_s=0$, we will have $Q_\pi(s,a) = \lambda$ for all $a\in\mathcal{A}$, which is in contradiction to the real situation that $Q_\pi(s,a)$ denotes the action values for different actions. Then we get the following soft-max relationship between the action-value function and the policy:
\begin{equation}\label{equ:first}
    \pi(a|s) = Z_s e^{\frac{Q_\pi(s,a)}{\mu_s}},
\end{equation}
where $Z_s = e^{-\frac{\lambda+\mu_s}{\mu_s}}$. Substituting \eqref{equ:first} into the third and the fifth equalities in Equation \eqref{equ:dual}, we have:
\begin{equation}\label{equ:prob_entropy}
\begin{aligned}
\sum_{a\in\mathcal{A}}Z_s e^{\frac{Q_\pi(s,a)}{\mu_s}} = 1 \\
-\sum_{a\in \mathcal{A}} Z_s e^{\frac{Q_\pi(s,a)}{\mu_s}} \log Z_s e^{\frac{Q_\pi(s,a)}{\mu_s}}=C_s,
\end{aligned}
\end{equation}
Substituting the first equation in Equation \eqref{equ:prob_entropy} into the second one, we have:
\begin{equation}\label{equ:entropy}
    -\sum_{a\in \mathcal{A}} p_s(a)\log p_s(a)=C_s,
\end{equation}
where
\begin{equation}\label{equ:soft_prob}
    p_s(a) = \frac{e^{\frac{Q_\pi(s,a)}{\mu_s}}}{e^{\sum_{a\in \mathcal{A}}{\frac{Q_\pi(s,a)}{\mu_s}}}}.
\end{equation}
Equations \eqref{equ:entropy} and \eqref{equ:soft_prob} precisely encode that $p_s(a)$ is a probability distribution that takes a soft-max form and its entropy equals to $C_s$. Then it is apparent that as $C_s$ approaches zero (i.e., the policy tends to become deterministic), $\mu_s$ approaches zero (since the limit of soft-max function is the arg-max function). In this situation, $\lambda$ must be larger than zero and thus the constant $Z_s$ equals to $1/e$.

\section{Adversarial training with value attack and PPO}\label{appendixB}
The objective of optimizing the perturbed policy $\pi_\theta(a|s+\delta_s)$ using PPO is:
\begin{equation}\label{equ:ppo_target_perturbed}
    L_\delta^{CLIP}(\theta) = \mathbb{E}_{t}\left[ \min\left(r_t(\theta) \hat A_{\delta}(s_t, a_t), \mathrm{clip}\left(r_t(\theta), 1-\rho, 1+\rho\right) \hat A_{\delta}(s_t, a_t) \right) \right],
\end{equation}
where $r^\delta(\theta) = \frac{\pi_\theta(a_t|s_t + \delta_{s_t})}{\pi_{\theta_{\mathrm{old}}}(a_t|s_t + \delta_{s_t})}$, and $\hat A_\delta(s,a)$ is an estimate of the advantage function $A_{\pi_\theta^\delta}(s,a)$ of the perturbed policy.

In practice, $\hat A_\delta(s,a)$ is estimated by the generalized advantage estimation (GAE) method (\citet{schulman2015high}). The pseudocode of ATPA implemented with PPO is summarized in Algorithm \ref{alg:defense_critic_ppo}. Our code is based on OpenAI baselines (\citet{baselines}) and uses the same hyperparameter configuration. The detailed hyperparameter configuration appears in Table \ref{tab:parameters}.

\begin{algorithm}
\caption{Adversarial Training with Policy Attack and PPO}
\label{alg:defense_critic_ppo}
\begin{algorithmic}[1]
\State Randomly initialize policy network $\pi_\theta(a|s)$, value network $V_\omega(s)$ of the attacked policy $\pi_{\theta}(a|s+\delta_s)$ with weights $\theta$, $\omega$
\State Set adversarial attack magnitude $\epsilon$, PGD steps $n$ and step size $\alpha$
\For {$\mathrm{iteration}=1, \cdots$}
    \For{$\mathrm{actor}=1,2,\cdots, N$}
    \State Rollout a trajectory $\tau_{\pi_\theta^\delta}=(s_0, a_0, r_0, \cdots, s_{T-1}, a_{T-1}, r_{T-1})$ by running $\pi_\theta(a|s+\delta_s)$ for $T$ time steps, where $\delta_s$ is obtained by running $\textsc{Attack}(\pi_\theta(a|s), \epsilon, n, \alpha)$
    \State Update value network weights $\omega$ using trajectory $\tau_{\pi_{\theta}^{\delta}}$
    \State Update policy parameter $\theta$ by maximizing Equation \eqref{equ:ppo_target_perturbed}
    \EndFor
\EndFor
\end{algorithmic}
\end{algorithm}

\begin{table}
  \caption{Hyperparameters configuration. $\beta$ is linearly annealed from 1 to 0 over the course of learning}\label{tab:parameters}
  \centering
  \begin{tabular}{l|l}
    \toprule
    Hyperparameter     & Value \\
    \hline
    Horizon ($T$)  & $128$     \\
    Num. epochs & $4$      \\
    Parallel actors ($N$) & $16$ \\
    Minibatch size & $32\times16$ \\
    Discount factor ($\gamma$) & $0.99$ \\
    GAE parameter ($\lambda$) & $0.95$ \\
    Clipping parameter ($\rho$) & $0.1\times \beta$ \\
    Learning rate & $2.5\times 10^{-4} \times \beta$ \\
    $\ell_\infty$ constraint ($\epsilon$) & $4.0$ or $2.0$ \\
    PGD step size ($\alpha$) & $2.0$ \\
    PGD steps ($n$) & $7$ \\
    Num. training steps & $10^8$  \\
    \bottomrule
  \end{tabular}
\end{table}

\FloatBarrier
\section{Some Figures}\label{appendixC}

\begin{figure}
\centering
\subfigure[Standard; Middle]{
\includegraphics[width=3.35cm]{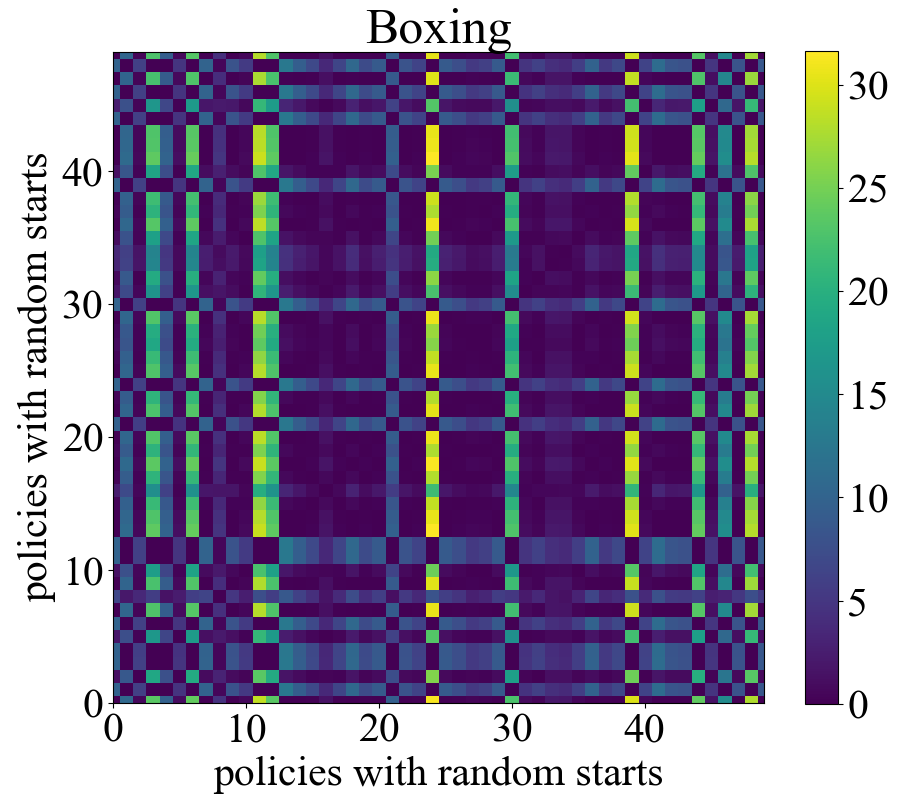}}%
\subfigure[Standard; End]{
\includegraphics[width=3.35cm]{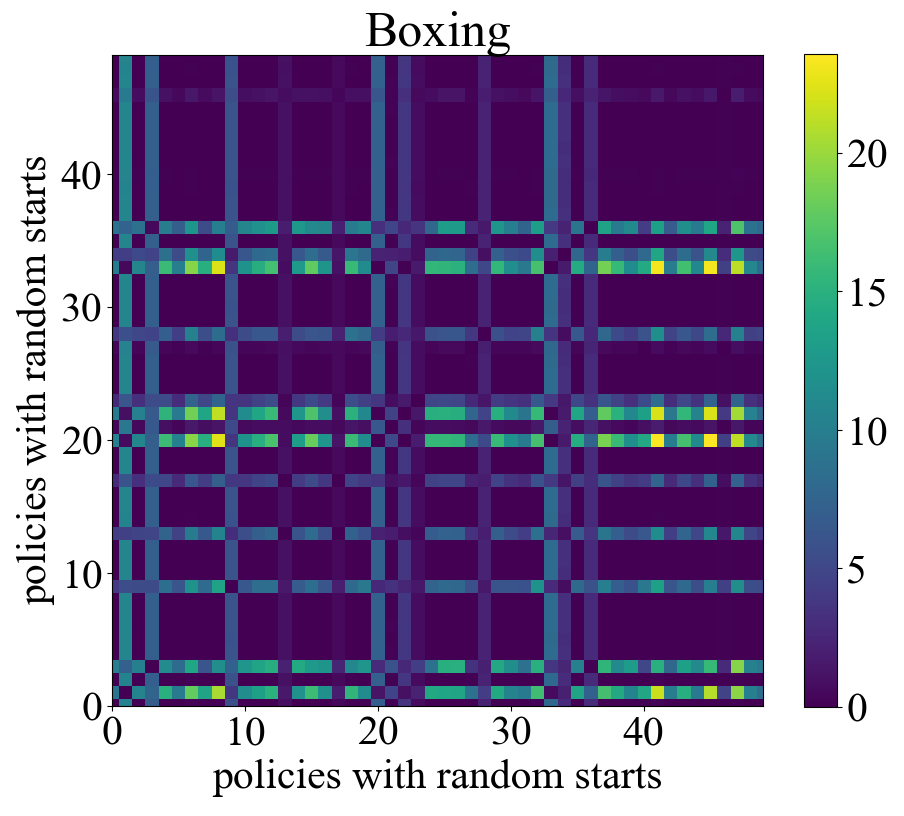}}%
\subfigure[ATPA; Middle]{
\includegraphics[width=3.35cm]{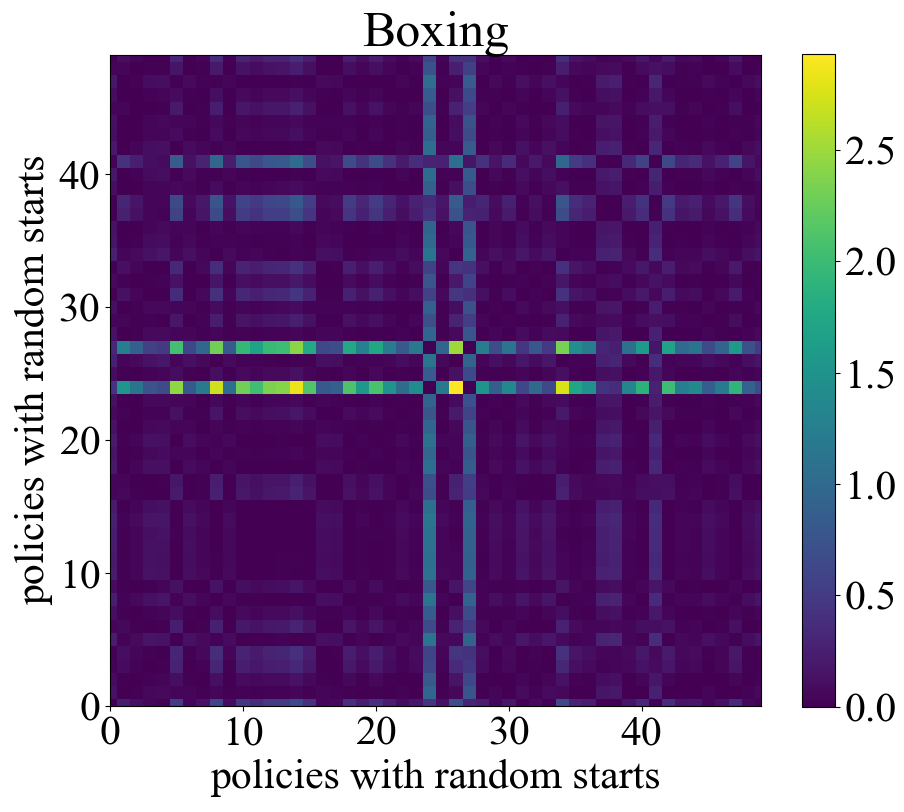}}%
\subfigure[ATPA; End]{
\includegraphics[width=3.5cm]{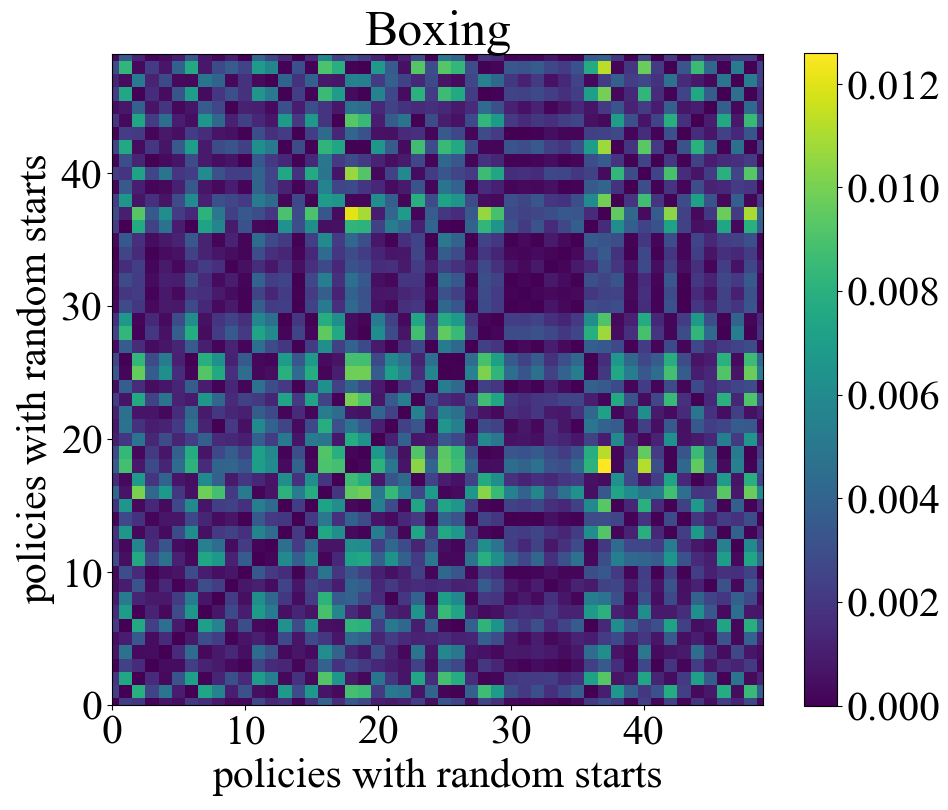}}%
\caption{KL divergence maps of perturbed policies (at different training stages (Middle, End)) with different initial points. Each pixel in the figures denotes the KL divergence value of two perturbed policies given by maximizing Equation \eqref{equ:policy_attack} with uniformly random initial points for $7$ projected gradient descent steps. Note that the maps are not symmetric because KL divergence is an asymmetric measure.}\label{fig:divergence_map}
\end{figure}

\begin{figure}
\centering
\subfigure[Boxing]{
\includegraphics[width=6.8cm]{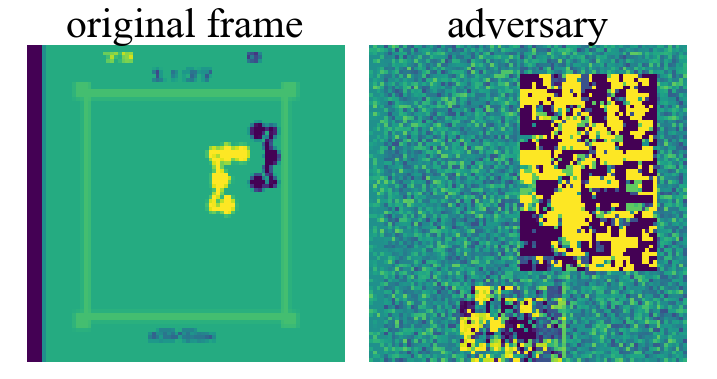}}%
\subfigure[Carnival]{
\includegraphics[width=6.8cm]{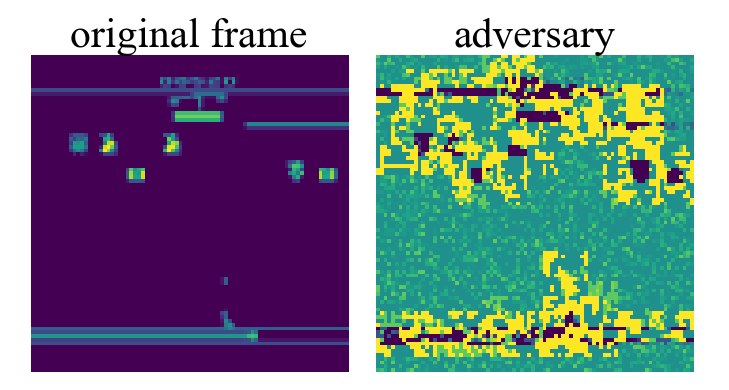}}%

\subfigure[SpaceInvaders]{
\includegraphics[width=6.8cm]{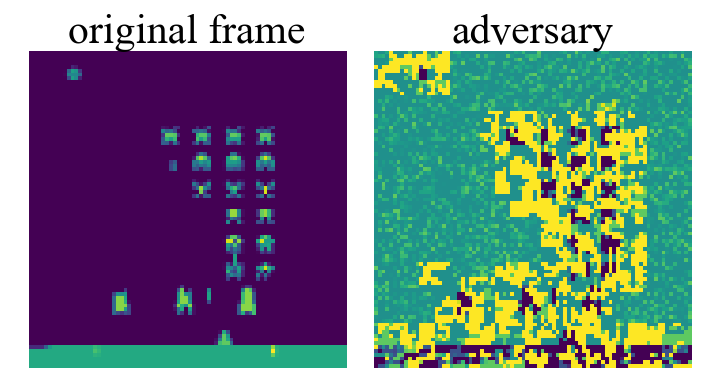}}%
\caption{Illustration of adversarial perturbations and the corresponding perturbation-free observations in different environments. The adversarial perturbations are given by maximizing Equation \eqref{equ:policy_attack} for $7$ projected gradient descent steps. Note that pixel values of original frames are within the range of $[0, 255]$ while those of perturbations are within the range of $[-4, 4]$.}\label{fig:adversarial_attacks}
\end{figure}

\end{document}